\documentclass[final]{X}

\usepackage{times}
\usepackage{epsfig}
\usepackage{graphicx}
\usepackage{amsmath}
\usepackage{amssymb}

\usepackage{multirow}
\usepackage{subfigure}
\usepackage{bm}
\usepackage{makecell}
\usepackage[dvipsnames]{xcolor}

\usepackage[pagebackref=true,breaklinks=true,colorlinks,bookmarks=false]{hyperref}

\begin{document}
	
	\title{Interactive Fusion of Multi-level Features for Compositional Activity Recognition}
	\author{Rui Yan$^1$ \quad
		Lingxi Xie$^2$ \quad
		Xiangbo Shu$^1$ \quad
		Jinhui Tang$^1$\thanks{Corresponding author}\\
		$^1$Nanjing University of Science and Technology, China \quad
		$^2$Huawei Inc.\\
		{\tt \small \{ruiyan,shuxb,jinhuitang\}@njust.edu.cn;198808xc@gmail.com}\\
	}
	\definecolor{mo_cyan}{rgb}{0, 1, 1}
	\definecolor{mo_red}{rgb}{1, 0, 0}
	\makeatletter
	\let\@oldmaketitle\@maketitle
	\renewcommand{\@maketitle}{\@oldmaketitle
		\includegraphics[width=1\linewidth]{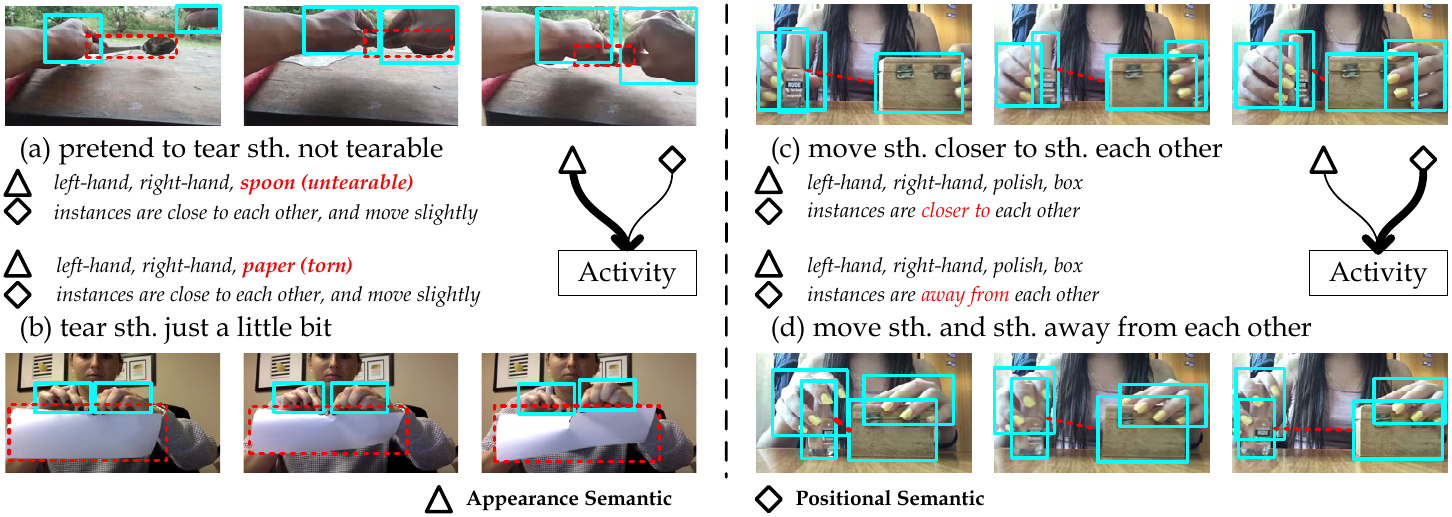}
		\refstepcounter{figure} Figure~\thefigure: Compositional examples in the Something-Else dataset~\cite{materzynska2020something}. We annotate all instances~(\textit{i.e.}, hands and objects) by the {\color{mo_cyan}cyan} boxes and highlight the major difference by the {\color{mo_red}red} dashed boxes or lines in each group of comparisons. By observing the appearance changes of objects~({\em e.g.}, \textit{\color{mo_red}untearable spoon} and \textit{\color{mo_red}torn paper}), human can easily distinguish between (a) and (b). 
		On the contrary, human understand ``closer to" in (c) and ``away from" in (d) by observing the relative displacement~(shown as the {\color{mo_red}red} dashed line) among instances, rather than objects' appearance. Motivated by this, we aim at fusing different types of information for understanding compositional actions. Best viewed in color.
		\label{fig:motivation}
		\bigskip\bigskip
	}
	\makeatother

\maketitle

\begin{abstract}
   To understand a complex action, multiple sources of information, including appearance, positional, and semantic features, need to be integrated. However, these features are difficult to be fused since they often differ significantly in modality and dimensionality. In this paper, we present a novel framework that accomplishes this goal by \textbf{interactive fusion}, namely, projecting features across different spaces and guiding it using an auxiliary prediction task. Specifically, we implement the framework in three steps, namely, positional-to-appearance feature extraction, semantic feature interaction, and semantic-to-positional prediction. We evaluate our approach on two action recognition datasets, Something-Something and Charades. Interactive fusion achieves consistent accuracy gain beyond off-the-shelf action recognition algorithms. In particular, on Something-Else, the compositional setting of Something-Something, interactive fusion reports a remarkable gain of $2.9\%$ in terms of top-1 accuracy.\footnote{The source code is available at \url{https://github.com/ruiyan1995/Interactive_Fusion_for_CAR}}
\end{abstract}

	\section{Introduction}
{Why can humans easily understand a complex action (\textit{e.g.}, ``\textit{Moving sth. and sth. closer to each other}" as shown in Figure~\ref{fig:motivation}(c)) even when it is performed with different objects in various environments?} Their ability to perform compositional reasoning between entities in the natural world is probably the most plausible explanation~\cite{spelke2007core}. For example, humans often first decompose the scene of Figure~\ref{fig:motivation}~(c) into some semantic instances~(\textit{i.e.}, two boxes and two hands), and then recognize the activity by observing the relative change of the distance between instances. This ability helps humans to learn some knowledge that is easy to be generalized to the novel environment with unseen combinations. Inspired by this, we hope to make machines show a similar capability in action recognition, namely, compositional action recognition~\cite{materzynska2020something,ji2020action}.

In the era of deep learning, most methods~\cite{carreira2017quo,lin2019tsm,yan2020social} extract appearance representations by strong backbones for action recognition. Due to the complexity of actions and the limited number of samples, the appearance representation will unavoidably bring inductive bias~\cite{torralba2011unbiased}~({\em e.g.}, the appearance-based model may confuse the actions in Figure~\ref{fig:motivation}~(c) and (d) which have almost the same appearance). Compositional action recognition just requires the model to be partly insensitive to the appearance of objects and scenes. Therefore, it is necessary for this task to extract features from other inputs~(\textit{e.g.}, positional information of instances that can be used to easily distinguish Figure~\ref{fig:motivation}~(c) from (d)) rather than the only image. Obviously, the positional feature does not work on some actions with the changes in the property of objects, \textit{e.g.}, the paper is torn by hands in Figure~\ref{fig:motivation}~(b). Therefore, fusing multi-level features becomes the main issue of compositional action recognition.

However, the features from different levels can vary significantly in modality and dimensionality. In general, appearance features extracted from images have thousands of dimensions, but positional features can even be described by vectors of $4$ dimensions.
Therefore, it is not appropriate to roughly merge multi-level features in a late fusion way, {\textit{e.g.},}~\cite{materzynska2020something} directly concatenates appearance and non-appearance features without any interaction between them.

{In this work, we integrate multi-level features in an \textbf{\textit{interactive}} way, \textit{i.e.}, projecting features across different spaces and guiding this process by an auxiliary task that does not require additional supervision.} Our approach includes the following three steps. i), \textbf{Positional-to-Appearance Feature Extraction}: builds instance-centric appearance representation from images according to tracklets and combine them with non-appearance features~\cite{materzynska2020something} into instance-centric joint representations; ii), \textbf{Semantic Feature Interaction}: builds category-aware pairwise relationships among these joint representations in the latent space to generate semantic features for each instance; iii), \textbf{Semantic-to-Positional Prediction}: projects the semantic features back to positional space by an auxiliary task. Predicting the future state of the instances is a good choice which not only does not require additional annotations but also helps the model understand the temporal changes of instances. Thus, we predict the location and offset of each instance to promote the interactive fusion of multi-level features.

The proposed approach is evaluated on two challenging datasets including the Something-Else dataset~\cite{materzynska2020something} and a fine-grained version of Charades~\cite{sigurdsson2016hollywood} annotated by~\cite{ji2020action}. Experimental results show that our model significantly outperforms state-of-the-art methods~\cite{materzynska2020something,wang2018videos,wang2018non} on compositional action recognition. In particular, on the Something-Else dataset, our approach achieves $2.9\%$ and $3.3\%$ boost on top-1 and top-5 accuracy, respectively, compared with the end-to-end model proposed in~\cite{materzynska2020something}. Besides, our model shows good generalization ability in the few-shot setting, implying that a better way of feature fusion is helpful to resist over-fitting.

	\section{Related Work}
\noindent{\bf Compositionality in Activity Recognition.}~Human activities are composed of a set of subactions in the temporal domain~\cite{shao2020intra,shao2020finegym}, and various subjects and objects in the spatial domain~\cite{ji2020action,baradel2018object}. Recent works proposed the compositional annotations or settings based on some popular video-based datasets~\cite{goyal2017something, sigurdsson2016hollywood}. For example, Materzynska~\textit{et al.}~\cite{materzynska2020something} provided object bounding box annotations for the	Something-Something dataset~\cite{goyal2017something} and presented a compositional setting in which there is no overlap between the verb-noun combinations in the training and testing set. Besides, built upon Charades~\cite{sigurdsson2016hollywood}, Action Genome~\cite{ji2020action} decomposed activities into spatiotemporal scene graphs as the intermediate representation for understanding them. To control the scene and object bias, Girdhar~\textit{et al.}~\cite{girdhar2019cater} created a synthetic video dataset, CATER, in which the events are broken up into several atomic actions in the spatial and temporal domain. In this work, we focus on generalizing compositional action to novel environments by integrating different levels of features in an interactive way.

\noindent{\bf Instance-centric Video Representation.}~For video understanding, previous works~\cite{wang2018non,lin2019tsm,carreira2017quo,ji20123d,feichtenhofer2019slowfast} focused on designing deep and powerful backbones to extract appearance features from each frame, however, in which rich relationships~\cite{ji2020action,krishna2017visual,ma2018attend,wang2018videos} between different instances~(objects or hands or persons, etc.) are hard to be mined in the spatial and temporal domain. To this end, some recent researches~\cite{baradel2018object,wang2018videos,ma2018attend,yan2020higcin,sun2018actor} focused towards extracting instance-centric representations from videos and building the spatial and temporal relationships among them. In this paper, we construct instance-centric representations from both low-level and high-level information for compositional action recognition.

\noindent{\bf Prediction in Video.}~{Predicting small image patches~\cite{qi2020learning, ranzato2014video,srivastava2015unsupervised} or full frames~\cite{denton2017unsupervised,mathieu2015deep,oh2015action} in videos has received increasing attention in recent years. A line of recent works~\cite{hsieh2018learning,kitani2012activity,huang2015approximate,ye2018interpretable,wu2016physics} focused on disentangling each frame into several instances and predict the state of them as mass, location, velocity, \textit{etc}. Sequence prediction mechanism has also been used in proxy tasks~\cite{han2019video,han2020memory,oord2018representation} to learn self-supervised representation for video. In this work, we predict the location and offset of instances from instance-centric semantic representations for promoting the fusion of multi-level features.}

\begin{figure*}[t]
	\begin{center}
		\includegraphics[width=1\linewidth]{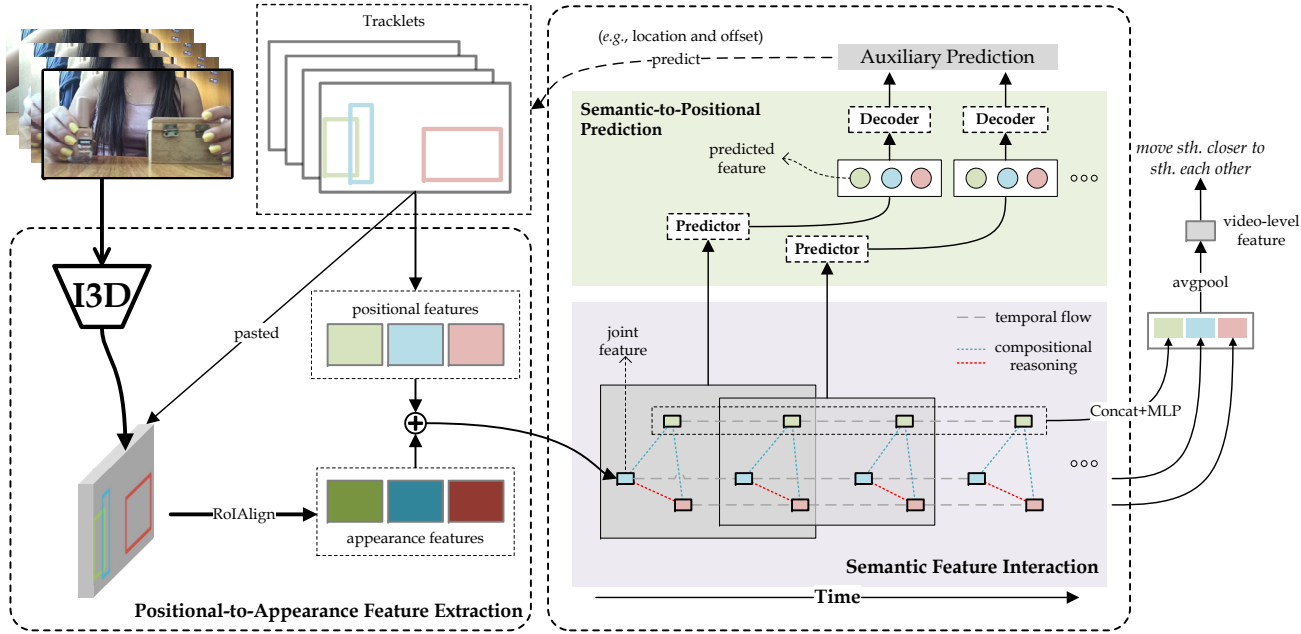}
	\end{center}
	\caption{Overview of the proposed framework. It takes $T$ frames sampled from a video, and the associated positional and category information of each instance in the scene as inputs. This framework is composed of three steps as follows. {\bf Positional-to-Appearance Feature Extraction}~(in Section~\ref{PAFE}): pastes a set of boxes on the image to extract instance-centric appearance features and concatenate them with positional features into basic joint representation; {\bf Semantic Feature Interaction}~(in Section~\ref{SFI}): {builds category-aware pairwise relationships among joint representations with the consideration of category information to achieve semantic features for each instance}; {\bf Semantic-to-Positional Prediction}~(in Section~\ref{SPP}): recovers part of high-level information from semantic representations by predicting the future positional information~(\textit{e.g.}, location and offset) of each instance. Best viewed in color.
	}
	\label{fig:overview}
\end{figure*}

\section{Our Approach}
\subsection{Problem Statement}
Formally, given a video with $T$ frames, $N$ instances~(\textit{e.g.}, objects and hands/persons), and the associated tracklets, we denote the RGB input of this video as $\bm{I} \in \mathbb{R}^{T \times H \times W \times 3}$, the tracklets of instances as $\bm{B} \in \mathbb{R}^{T \times N \times 4}$, the category information of instances as $\bm{C} \in \mathbb{R}^{T \times N \times 1}$~(\textit{only indicates ``hand/person" and ``object"}), and the activity label of this video as $l$. Compositional activity recognition aims at understanding the unseen combination of action~(performed by subjects including hands and persons) and objects in each video. 

Based on the above inputs, there are multiple levels of features can be used to understand a complex activity. Such as low-level appearance features extracted from images $\bm I$ that contain the attributes of instances or the environment. Beyond that, we can also obtain high-level positional features that describe the trajectories~(\textit{i.e.}, motions) of instances from tracklets $\bm B$, and the high-level category feature used to identify each instance from $\bm C$. In general, appearance features take on thousands of dimensions, but non-appearance~(\textit{e.g.}, positional and category) features can be represented by vectors of several tens of dimensions.

\cite{materzynska2020something} has validated that both low-level and high-level features are useful for understanding a complex compositional activity. But, the features extracted from $\bm{B}$, $\bm{C}$, and $\bm{I}$ were simply fused in a late stage. This pipeline ignored that multi-level features often significantly differ in modality and dimensionality.

\subsection{Interactive Fusion of Multi-level Features}
To this end, we propose a simple yet effective interactive fusion framework that leverages high-level information to assist the feature extraction from low-level information and predict high-level future information with an auxiliary loss. The general formulation of this framework can be abstracted as:
\begin{equation}
	\begin{aligned}\label{eq::general_framework}
		{\bm G} =& \mathcal{G}(\mathcal{F}({\bm B}, {\bm A}), \bm{C}),\\
		\mathcal{L} =& \mathcal{L}_\mathrm{reg}(\bm{G}, {l})+\lambda\mathcal{L}_\mathrm{aux}({\mathcal H}(\bm{G}), \bm{B}),
	\end{aligned}
\end{equation}where $\bm{A}$ denotes the low-level appearance feature extracted from images~$\bm I$; $\bm{B}$ and $\bm{C}$ represent the high-level positional and category information, respectively; $l$ is the activity label of each video. The total loss of this framework is split into the main recognition loss $\mathcal{L}_\mathrm{reg}$ and the auxiliary loss $\mathcal{L}_\mathrm{aux}$, and $\lambda$ is used to balance them. The framework has three major components. \textbf{First}, we \textit{project} $\bm{B}$ to fit $\bm{A}$ in a high-dimension feature space and extracts the basic instance-centric joint features by $\mathcal{F}(\bm{B}, \bm{A})$; \textbf{Second}, we further explore the semantic interaction among them by $\mathcal{G}(\cdot)$; \textbf{Third}, we employ $\mathcal{H}(\cdot)$ to project semantic representations from the latent space back to the low-dimensional space of $\bm{B}$ supervised by an auxiliary loss $\mathcal{L}_\mathrm{aux}$. Within such a generic framework, we complete the feature fusion through performing multiple interactions between different levels of features. 

{Note that $\mathcal{F}(\cdot)$, $\mathcal{G}(\cdot)$, and $\mathcal{L}_\mathrm{aux}$ can be designed into different forms with the requirement that $\mathcal{F}(\cdot)$ is built on instances rather than frames, $\mathcal{G}(\cdot)$ needs further leverage more existing supervision to promote feature fusion in the latent space, and $\mathcal{L}_\mathrm{aux}$ can be free from the additional annotation.

}

\subsubsection{Positional-to-Appearance Feature Extraction}\label{PAFE}
{Function $\mathcal{F}(\cdot)$ aims at mixing positional and appearance information during the feature extraction stage by {\em projecting} a set of boxes on an image. In this work, we adopt a simple way: extracting appearance representation for each instance from the image according to the high-level positional information. This method makes full use of positional supervision to eliminate the scene bias in representations.}

Specifically, we sample $T$ frames from each video and employ I3D~\cite{carreira2017quo} to extract the spatiotemporal appearance representation from images $\bm{I}$. The output of the last convolutional layer is a feature map with the dimensions of $\frac{T}{2} \times H \times W \times d_{\mathrm{fea}}$, where $d_{\mathrm{fea}}$ is the channel number. To get $N$ instance-centric features, we apply a RoIAlign~\cite{girshick2015fast} on the top of the last convolutional layer to crop and rescale the appearance feature for each instance according to the associated tracklets $\bm{B}$. Finally, we denote the appearance features as ${\bm F}_{\mathrm{app}} \in \mathbb{R}^{\frac{T}{2} \times N \times d_{\mathrm{app}}}$. 

Furthermore, we combine the appearance features with non-appearance features used in~\cite{materzynska2020something}. We can easily achieve positional features ${\bm F}_{\mathrm{bb}} \in \mathbb{R}^{\frac{T}{2} \times N \times d_{\mathrm{bb}}}$ from a quadruple of ${\bm p}^{t}_{i} = \left\langle c_{\mathrm{x}}, c_{\mathrm{y}}, w, h \right\rangle $~(where $c_{\mathrm{x}}$ and $c_{\mathrm{y}}$ are the center coordinate, and $w$ and $h$ represent its width and height) and category features ${\bm F}_{\mathrm{cate}} \in \mathbb{R}^{\frac{T}{2} \times N \times d_{\mathrm{cate}}}$ from category information $\bm{C}$ by word embedding. Then, ${\bm F}_{\mathrm{bb}}$ and ${\bm F}_{\mathrm{cate}}$ are concatenated and embedded into non-appearance representation ${\bm F}_{\mathrm{non\_app}} \in \mathbb{R}^{\frac{T}{2} \times N \times d_{\mathrm{non\_app}}}$ via an MLP. Finally, we concatenate ${\bm F}_{\mathrm{app}}$ and ${\bm F}_{\mathrm{non\_app}}$ at the last dimension as basic instance-centric joint representations $\bm F \in \mathbb{R}^{\frac{T}{2} \times N \times d_{\mathrm{joint}}}$ for each video, where $d_{\mathrm{joint}}=(d_{\mathrm{app}} + d_{\mathrm{non\_app}})$.

\subsubsection{Semantic Feature Interaction}\label{SFI}
{As the core elements of our framework, basic instance-centric joint representations extracted from each instance are isolated now, which limits the fusion of different levels of information. To this end, we instantiate function $\mathcal{G}(\cdot)$ by constructing pairwise relationships between each instance with the consideration of category information. Details are presented as follows.}

\noindent{\bf Compositional reasoning.}~Different from previous relational modules~\cite{materzynska2020something,wang2018non,yan2020higcin} that construct relationships between a specific feature node and its neighbors indistinguishably, we propose a {\em compositional reasoning} to model the different types of pairwise relationships between each instance with the consideration of their category information. Formally, we represent each video by a set of instance-centric joint features as $\bm F \in \mathbb{R}^{\frac{T}{2} \times N \times d_{\mathrm{joint}}}$ and perform compositional reasoning among them as:
\begin{equation}
	\begin{aligned}\label{eq::compostional_reasoning}
		{\bm g}_{i}^{t} &= \mathcal{G}({\bm F}^{t}, {\bm C}^{t}) = \psi^\mathrm{S} \Big({\bm f}_{i}^{t} + \!\! \sum_{\forall (i,j) \in \mathcal{E}_{\mathrm{ss}}}\!\!{\phi}_\mathrm{ss}([{\bm f}_{i}^{t}, {\bm f}_{j}^{t}])\\
		&+ \!\! \sum_{\forall (i,j) \in \mathcal{E}_\mathrm{so}}\!\!{\phi}_\mathrm{so}([{\bm f}_{i}^{t}, {\bm f}_{j}^{t}]) + \!\! \sum_{\forall (i,j) \in \mathcal{E}_\mathrm{oo}}\!\!{\phi}_\mathrm{oo}([{\bm f}_{i}^{t}, {\bm f}_{j}^{t}])
		\Big),
	\end{aligned}
\end{equation}
where ${\bm f}_{i}^{t}$ and ${\bm g}_{i}^{t}$ denote the basic joint and semantic features for $i$-th instance at $t$-th time step. $[\cdot, \cdot]$ represents a concatenation operation that used to compose a pair of instance-centric features, \textit{i.e.}, ${\bm f}_{i}^{t}$ and ${\bm f}_{j}^{t}$ where $j \neq i$. $\mathcal{E}_\mathrm{ss}$, $\mathcal{E}_\mathrm{so}$, and $\mathcal{E}_\mathrm{oo}$ represent different instance-pair sets~(\textit{i.e.}, subject-subject, subject-object, and object-object interactions, where subject indicates hand or person) that can be defined according to category information $\bm{C}$. Functions ${\phi}_\mathrm{ss}(\cdot)$, ${\phi}_\mathrm{so}(\cdot)$, and ${\phi}_\mathrm{oo}(\cdot)$ designed to encode different pairwise interactions and $\psi^\mathrm{S}(\cdot)$ used to fuse them with basic joint features, can be implemented by MLPs. This module outputs instance-centric semantic features ${\bm G} \in \mathbb{R}^{\frac{T}{2} \times N \times d_\mathrm{sem}}$.

\noindent{\bf Temporal flow.}~Given semantic features ${\bm G}$, we further aggregate them for each instance over time. To be flexible, we apply RNN~\cite{williams1989learning} architecture to capture the temporal dependencies as:
\begin{equation}\label{eq::temporal_flow}
	{\bm Z}_{i} = \psi^\mathrm{T}(\mathtt{Concat}(\mathtt{RNN}([{\bm g}_{i}^{1}, {\bm g}_{i}^{2}, \cdots, {\bm g}_{i}^{\frac{T}{2}}]; {\bm \theta}^\mathrm{\tau}))),
\end{equation}where ${\bm Z}_{i}$ is the spatiotemporal representation of the $i$-th instance, ${\bm \theta}^\mathrm{\tau}$ denotes the learnable parameters used in ${\mathtt{RNN}}$, $\mathtt{Concat}(\cdot)$ is used to concatenate all $\frac{T}{2}$ outputs from ${\mathtt{RNN}}$, and $\psi^\mathrm{T}(\cdot)$ is an MLP used to encode the concatenated features. $\bm Z = \{\bm Z_{0}, \bm Z_{1}, \bm Z_{2}, \cdots, \bm Z_{N}\}$ is average pooled as the final video-level representation used to predict the compositional activity. 

\subsubsection{Semantic-to-Positional Prediction}\label{SPP}
{To promote the fusion between appearance and non-appearance features, we implement $\mathcal{H}(\cdot)$ and $\mathcal{L_\mathrm{aux}}$ by predicting the future positional information of each instance from the observed semantic features. Instead of only predicting the accurate coordinates of each instance, we also estimate the offsets of it. Details are described as follows.
}

\noindent{\bf Predictor}. We first estimate the future state of each instance by the observed features. Formally, given the previous $t$ observed semantic features of $i$-th instance as $\{ {\bm g}^{0}_{i}, {\bm g}^{1}_{i}, \cdots, {\bm g}^{t}_{i} \}$, we predict the $(t+1)$-th state as:
\begin{equation}
	{\hat {\bm g}}^{t+1}_{i} = \mathcal{H}(\bm G) = {\mathtt{RNN}}(\{{\bm g}^{0}_{i}, {\bm g}^{1}_{i}, \cdots, {\bm g}^{t}_{i} \}; {\bm \theta}^\mathrm{\tau}),\label{eq::predictor}
\end{equation}Here, we denote $T_\mathrm{obs}$ as the number of frames observed by the {predictor} in each step, thus $t \in [T_\mathrm{obs}:\frac{T}{2}-1]$. $\mathtt{RNN}$ is used to modeling the temporal structure hidden in sequence data and the last output is viewed as the predicted state ${\hat {\bm g}}^{t+1}_{i}$. More importantly, recurrent architecture do not sensitive to the length of input data, which is flexible for modeling the temporal information on sequences of variable length. Therefore, this predictor can share the parameters ${\bm \theta}^\mathrm{\tau}$ with the recurrent architecture used in Eq.~\ref{eq::temporal_flow}.

\noindent{\bf Decoder.} Given the predicted state ${\hat {\bm g}}^{t+1}_{i} \in \mathbb{R}^{d_\mathrm{sem}}$ for the $i$-th instance in time $t+1$, we first estimate its $4$-dimension positional information via a simple linear layer as $\hat{\bm p}^{t+1}_{i} = {\bm W}_{p}^\top{\hat {\bm g}}^{t+1}_{i}$, where ${\bm W}_{p} \in \mathbb{R}^{d_\mathrm{sem} \times 4}$. Besides, we also estimate the offsets~(\textit{i.e.}, the difference between the center coordinates~$<c_\mathrm{x}, c_\mathrm{y}>$ of the same instance in two consecutive time) of each instance as $\hat{\bm o}^{t+1}_{i} = {\bm W}_{o}^\top{\hat {\bm g}}^{t+1}_{i}$, where ${\bm W}_{o} \in \mathbb{R}^{d_\mathrm{sem} \times 2}$.

\noindent{\bf Auxiliary loss.} For simplicity, we measure the error between prediction and ground truth by euclidean distance and sum up them over space and time as:
\begin{equation}\label{pred_loss}
	{\mathcal{L}_\mathrm{aux}} = \sum_{\forall t} \sum_{i=1}^{N} ({\| \hat{\bm p}^{t+1}_{i}-{\bm p}^{t+1}_{i} \|}^{2}_{2} + {\| {{\hat{\bm o}}}^{t+1}_{i}-{\bm o}^{t+1}_{i} \|}^{2}_{2}),
\end{equation}where $t \in [T_\mathrm{obs}:\frac{T}{2}-1]$ and $N$ is the number of instances in the video. ${\bm p}^{t+1}_{i}$ and ${\bm o}^{t+1}_{i}$ are ground truth position and offset, respectively. ${\bm o}^{t+1}_{i}=({\bm p}^{t+1}_{i}-{\bm p}^{t}_{i})[:2]$ where $[:2]$ takes only previous two elements of vectors.

\subsection{Implementation Details}
\noindent{\bf Input}: Our approach takes as input $T$ frames uniformly sampled from each video, and the associated tracklets and category information for $N(=4)$~semantic instances~(hands/persons and objects, no more than four) in the scene. Each frame is resized into the resolution of $224 \times 224$. 

\noindent{\bf Network Architecture}: The frame-based feature extractor is an important component for action recognition. For a fair comparison, we only employ the I3D model~\cite{carreira2017quo} built on ResNet-50 as the backbone of our approach and more details are introduced in~\cite{wang2018videos}. The RoIAlign~\cite{girshick2015fast} extracts region-based features for each instance with the size of $3 \times 3$ on top of the last convolutional layer. Thus, the RoIAlign layer generates a $N \times 3 \times 3 \times d_\mathrm{fea}$ output features for each video which is then flatten and embedded into $N \times d_\mathrm{app}$ via a linear transformation, where $N=4$ and $d_\mathrm{fea}=d_\mathrm{app}=512$. 

Besides, the MLP used to fuse non-appearance features and $\psi^\mathrm{T}(\cdot)$ used in the temporal flow, are composed of two linear layers with the output dimension of $512$, thus $d_\mathrm{non\_app}=512$ and $d_\mathrm{joint} = 1024$. ${\phi}_\mathrm{ss}(\cdot)$, ${\phi}_\mathrm{so}(\cdot)$, ${\phi}_\mathrm{oo}(\cdot)$ and $\psi^\mathrm{S}(\cdot)$ defined in Eq.~\ref{eq::compostional_reasoning} are implemented by MLPs each of which is single linear layer with the output dimension of $1024$. The RNN used in Eq.~\ref{eq::temporal_flow} and Eq.~\ref{eq::predictor} is implemented by a LSTM~\cite{hochreiter1997long} with two layers and the dimension of hidden states is equal to $1024$. $T_\mathrm{obs}$ used in prediction is set to $\frac{T}{4}$. ${\mathcal{L}}_\mathrm{reg}$ is a simple cross entropy loss.

	\section{Experiments}

\subsection{Setup}
\noindent{\bf Dataset.}
We evaluate the proposed approach on two datasets: i), {\em Something-Else}~\cite{materzynska2020something}~(built on~\cite{goyal2017something} with the compositional setting {forcing the combinations of action and objects cannot overlap between training and testing sets}). Something-Else contains $174$ categories of activities but only $112,795$ videos~($54,919$ for training and $57,876$ for testing) with the compositional setting; ii), {\em Charades}~\cite{sigurdsson2016hollywood} with only 9,601 videos annotated by Action Genome~\cite{ji2020action}. We still follow the official train-validation split provided in~\cite{sigurdsson2016hollywood} and each video may contain multiple actions out of 157 classes.

In this work, we do not evaluate our approach on the detected bounding boxes mentioned in~\cite{materzynska2020something} as only the ground-truth of them is released in~\cite{ji2020action,materzynska2020something}. Certainly, it is another challenge for this task to be free from time-consuming ground-truth annotations, which is worth exploring in the future. 

\noindent{\bf Methods and baselines.}
{To figure out the effectiveness of different components designed in our approach, we compare the following methods:
	\begin{itemize}
		\item {\bf I3D}: {A frame-level baseline that apply a 3D ConvNet~\cite{carreira2017quo,wang2018videos} built on ResNet-50 to extract spatio-temporal representations.}
		\item {\bf Ours-base}: A basic version of our model that average pools the basic instance-centric joint representations in the spatial domain as frame-level representations and concatenate them over time followed by an MLP.
		\item {\bf Ours-SFI}: One variant of our approach that further explore the latent interactions among the basic joint representations by the block of Semantic Feature Interaction~(SFI).
		\item {\bf Ours-SFI + Pred}: The complete version of our approach that further apply semantic-to-positional prediction to promote the fusion between different types of features, based on the previous variant.
	\end{itemize}
} Moreover, we also compare our approach with recent state-of-the-art methods on object-based video understanding, such as NL~\cite{wang2018non}, STRG~\cite{wang2018videos}, and STIN~\cite{materzynska2020something}. Ours-NLs, Ours-STRG, and Ours-STIN represent that we build these modules on basic instance-centric joint representations under our framework. Notably, (I3D, Ours-SFI+Pred) and (I3d, STIN + NL) represent that combine the frame-level I3D model with the existing model ensemble but train them separately, all other methods are trained in an end-to-end fashion. 

\noindent{\bf Training details.}
We employ SGD with fixed hyper-parameters ($0.9$ momentum and  $10^{-4}$ weight decay) to optimize all our models with the initial learning rate $10^{-2}$. The I3D model used in this work is pre-trained on Kinetics-400~\cite{kay2017kinetics}.

\subsection{Results on Something-Else}
Similar to~\cite{materzynska2020something}, we train non-appearance based models~(that do not use any appearance information) in $50$ epochs and reduce the learning rate to $1/10$ of the previous one at epoch $35$ and $45$. However, we train appearance-based models in $30$ epochs and decay the learning rate by the factor of $10$ for every $5$ epochs, since appearance representation can help the model converge faster. All our models are fed with $8$ frames and trained in a batch-size of $72$ on this dataset. 

\noindent{\bf Ablation study.}~We first conduct some ablation experiments to evaluate each component of our approach using the ground-truth bounding boxes, and the results are reported in Table~\ref{Ablation}. All these methods are the variants of our approach, except the I3D. 

As we can see that I3D performs poorly when using only $8$ frames, though its performance is reported as $46.8\%$ in terms of top-1 accuracy in~\cite{materzynska2020something} with $16$ frames. With the same setting, our basic interactive fusion framework, \textit{i.e.}, Ours-base~(appearance), improves the top-1 and top-5 accuracy by $16.1\%$ and $14.1\%$, compared with the frame-based model, \textit{i.e.}, I3D~($T=8$). Ours-base~(non-appearance) can easily surpass the I3D model, indicating the effectiveness of non-appearance features on this task. As expected, Ours-base achieves significant gains in accuracy benefited by fusing both non-appearance and non-appearance features.

Equipped with the interactions among semantic representations, Ours-SFI brings $2.2\%$ and $1.4\%$ improvements for the basic framework on top-1 and top-5, respectively. After that, Ours-SFI + Pred improves the top-1 and top-5 by $1.3\%$ and $1.0\%$ again by introducing auxiliary positional prediction. Finally, we combine Ours-SFI + Pred and I3D ensemble and achieve the best results on this dataset against other variants.

\begin{table}[t]
	\begin{center}
		\begin{tabular}{l|cc}
			\multirow{2}*{Method}  & \multicolumn{2}{c}{Accuracy~($\%$)} \\
			& top-1 & top-5\\
			\hline\hline
			I3D~($T=8$)~\cite{carreira2017quo,wang2018videos} & $40.0$ & $69.3$\\
			I3D$^{\dag}$~($T=16$)~\cite{materzynska2020something,carreira2017quo,wang2018videos} & $46.8$ & $72.2$\\
			\hline
			Ours-base~(appearance) & $54.8$ & $82.3$\\
			Ours-base~(non-appearance) & $44.1$ & $74.0$\\
			{Ours-base} & $56.1$ & $83.4$\\
			Ours-SFI & $58.3$ & $84.8$\\
			Ours-SFI + Pred & {$59.6$} & {$85.8$}\\
			I3D, Ours-SFI + Pred & $\bm{61.0}$ & $\bm{86.5}$\\
		\end{tabular}
	\end{center}
	\caption{Ablation study on Something-Else with the compositional setting.   {All our methods take as input only 8 frames and the appearance-based methods are built on I3D.}}\label{Ablation}
\end{table}

\begin{table}[t]
	\begin{center}
		\begin{tabular}{l|ll}
			\multirow{2}*{Method}  & \multicolumn{2}{c}{Accuracy~($\%$)} \\
			& top-1 & top-5\\
			\hline\hline
			I3D + STIN + NL & 51.7 & 80.5\\
			Ours-NLs~\cite{wang2018non} &$54.2$ &$81.4$ \\
			Ours-STRG~\cite{wang2018videos} &$56.4$&$83.6$ \\
			Ours-STIN~\cite{materzynska2020something} &$57.5$& $84.4$\\
			\hline\hline
			I3D$^{\dag}$ + STIN + NL~\cite{materzynska2020something} & $54.6$ & $79.4$\\
			Ours-SFI + Pred & ${59.6}^{\bm{+5.0}}$ & ${85.8}^{\bm{+6.4}}$\\
			\hline\hline
			I3D$^{\dag}$, STIN + NL~\cite{materzynska2020something} & $58.1$ & $83.2$\\
			I3D, Ours-SFI + Pred & $\bm{61.0}^{\bm{+2.9}}$ & $\bm{86.5}^{\bm {+3.3}}$\\
		\end{tabular}
	\end{center}
	\caption{Compositional action recognition on the Something-Else dataset. I3D$^{\dag}$ is pre-trained with 16 frames.}\label{tab:STHELSE_STOA}
\end{table}

\noindent{\bf Comparisons with the state-of-the-art method.} Besides, we also compare our approach with the most related work~\cite{materzynska2020something} and some reasoning modules~\cite{wang2018non,wang2018videos,materzynska2020something} that can be easily plugged into our approach, as shown in Table~\ref{tab:STHELSE_STOA}. Overall, Ours-SFI + Pred is superior to all existing methods and (I3D, Ours-SFI + Pred) achieves state-of-the-art performance on both top-1 and top-5 accuracy.

{In particular, Ours-SFI + Pred significantly improves the top-1 and top-5 accuracy by $5.0\%$ and $6.4\%$, respectively, compared with I3D + STIN + NL. In an ensemble way, (I3D, Ours-SFI + Pred) still gains $2.9\%$ and $3.3\%$ compared with~\cite{materzynska2020something}. Compared with NL~\cite{wang2018non}, STRG~\cite{wang2018videos} and STIN~\cite{materzynska2020something}, Ours-SFI + Pred has remarkable improvements in accuracy benefited from the auxiliary prediction task which promotes the fusion of multi-level features.

}

\begin{table}[h]
	\begin{center}
		\resizebox{\columnwidth}{!}{%
			\begin{tabular}{l|cc|ll}
				\multirow{2}*{Method}  & \multicolumn{2}{c|}{\em base} & \multicolumn{2}{c}{\em few shot}\\
				& top-1 & top-5 & 5-S & 10-S\\
				\hline\hline
				I3D~($T=8$)~\cite{carreira2017quo,wang2018videos}&$66.0$&$89.8$&$18.2$&$20.1$\\
				I3D$^{\dag}$~($T=16$)~\cite{materzynska2020something,carreira2017quo,wang2018videos}&$73.6$&$92.2$&$21.8$&$26.7$\\
				\hline
				{I3D$^{\dag}$ + STIN + NL~\cite{materzynska2020something}} &$80.6$ &$95.2$ &$28.1$ &$33.6$ \\
				Ours-base &$77.3$ &$94.4$ &$24.3$ & $29.8$\\
				Ours-NLs~\cite{wang2018non} &$76.6$ &$94.6$ &$25.3$ & $30.9$\\
				Ours-STRG~\cite{wang2018videos} &$77.9$&$95.0$&$26.6$&$32.4$ \\
				Ours-STIN~\cite{materzynska2020something} & $76.4$ & $94.1$ & $24.6$ &$29.4$\\
				Ours-SFI & $79.2$ & $95.5$ & $29.6$&$34.9$\\
				Ours-SFI + Pred & $79.5$ & $95.5$ & $30.7$ &$36.2$\\ 
				\hline\hline
				{I3D$^{\dag}$ + STIN + NL~\cite{materzynska2020something}} &$80.6$ &$95.2$ &$28.1$ &$33.6$ \\
				Ours-SFI + Pred & $79.5$ & $95.5$ & ${30.7}^{+\bm{2.6}}$ &${36.2}^{+\bm{2.6}}$\\ 
				\hline\hline
				\makecell[l]{I3D$^{\dag}$, STIN + NL~\cite{materzynska2020something}} &81.1 &96.0 &34.0 &40.6 \\
				I3D$^{\dag}$, Ours-SFI + Pred & ${\bm{84.1}}$ & $\bm{97.2}$&${\bm{35.3}^{+\bm{1.3}}}$&${\bm{41.7}^{+\bm{1.1}}}$\\
			\end{tabular}%
		}
	\end{center}
	\caption{Few-shot compositional action recognition on the Something-Else dataset. I3D$^{\dag}$ is trained/pre-trained with 16 frames.}\label{table::few_shot}
\end{table}

\noindent{\bf Few-shot setting.}~To further evaluate the generalization capability of our approach, we conduct some experiments on the few-shot setting which has 88 base and 86 novel categories proposed in~\cite{materzynska2020something}. For a fair comparison, we follow the same settings and training strategies used in~\cite{materzynska2020something}. We first train our approach on base categories and then directly finetune it on all the novel categories. More details can be found in~\cite{materzynska2020something}. In this work, only 5-shot and 10-shot results are reported in Table~\ref{table::few_shot}.

As expected, our approach shows strong generalization on the few-shot setting against other methods. In particular, Ours-SFI achieves the best results on 5/10-shots, compared with previous all joint-training methods~\cite{materzynska2020something,wang2018non,wang2018videos} and the frame-based method I3D~\cite{wang2018videos}. With the help of the auxiliary prediction branch, Ours-SFI + Pred further boosts the performance with $1.1\%$ and $1.3\%$ on 5-Shot and 10-Shot, respectively. Ours-SFI + Pred outperforms I3D$^{\dag}$ + STIN + NL whose I3D is pre-trained with $16$ frames by $2.6\%$ in both 5-shot and 10-shot setting, though trail it on the base categories. {Finally, we combine our approach and the I3D model in an ensemble way, and achieve state-of-the-art results.}

\subsection{Results on Charades}
There are multiple actions in each video of this dataset, the positional prediction of each instance is meaningless because no one knows which action will appear in the next frame. Therefore, we drop the auxiliary prediction branch from our approach and train all models in $50$ epochs and decay the learning rate to $1/10$ of the previous one at epoch $35$ and $45$. Similar to previous works~\cite{wang2018non,wang2018videos}, we combine a sigmoid layer and the binary cross-entropy loss to handle the multi-label classification problem and report the mean average precision (mAP) results in Table~\ref{tab:Charades}.

To avoid inaccurate detection, we take only $T=32$ sampled from annotated frames provided by~\cite{ji2020action} as the input of all methods on this dataset. Therefore, with just a part of the original sequence is observed, the baseline method I3D can only achieve $17.9\%$ mAP that is lower than its performance on the original dataset, but it does not affect the following comparisons. We can see that both I3D + STIN and I3D+ STIN + NL trail the I3D model on mAP but Ours-base improves it by $1.7\%$, suggesting that naive late fusion may harm the joint representation when some levels of features are useless. Equipped with our interactive fusion framework, Ours-STIN achieves a remarkable gain of $6.8\%$, compared with I3D + STIN~\cite{materzynska2020something}. As expected, Ours-SFI is superior to recent relational methods~(\textit{i.e.}, Ours-NLs/STRG/STIN), benefited by the compositional reasoning which considers the different types of interactions among instances.

\begin{table}[t]
	\begin{center}
		\begin{tabular}{l|c}
			{Method}  & {mAP~($\%$)} \\
			\hline\hline
			I3D~\cite{carreira2017quo,wang2018videos} & $17.4$ \\
			I3D + STIN~\cite{materzynska2020something,wang2018videos} & $14.4$ \\
			I3D + STIN + NL~\cite{materzynska2020something,wang2018videos} & $15.6$ \\
			Ours-base & $19.1$ \\
			Ours + NLs~\cite{wang2018non} & $19.1$ \\
			Ours + STRG~\cite{wang2018videos} & $21.7$ \\
			Ours + STIN~\cite{materzynska2020something} & $21.2$ \\
			Ours-SFI & $\bm{25.2}$ \\
		\end{tabular}
	\end{center}
	\caption{Action recognition on the Charades dataset.}~\label{tab:Charades}
\end{table}

\subsection{Diagnostic Studies}
\begin{figure}[h]
	\begin{center}
		\includegraphics[width=0.85\linewidth]{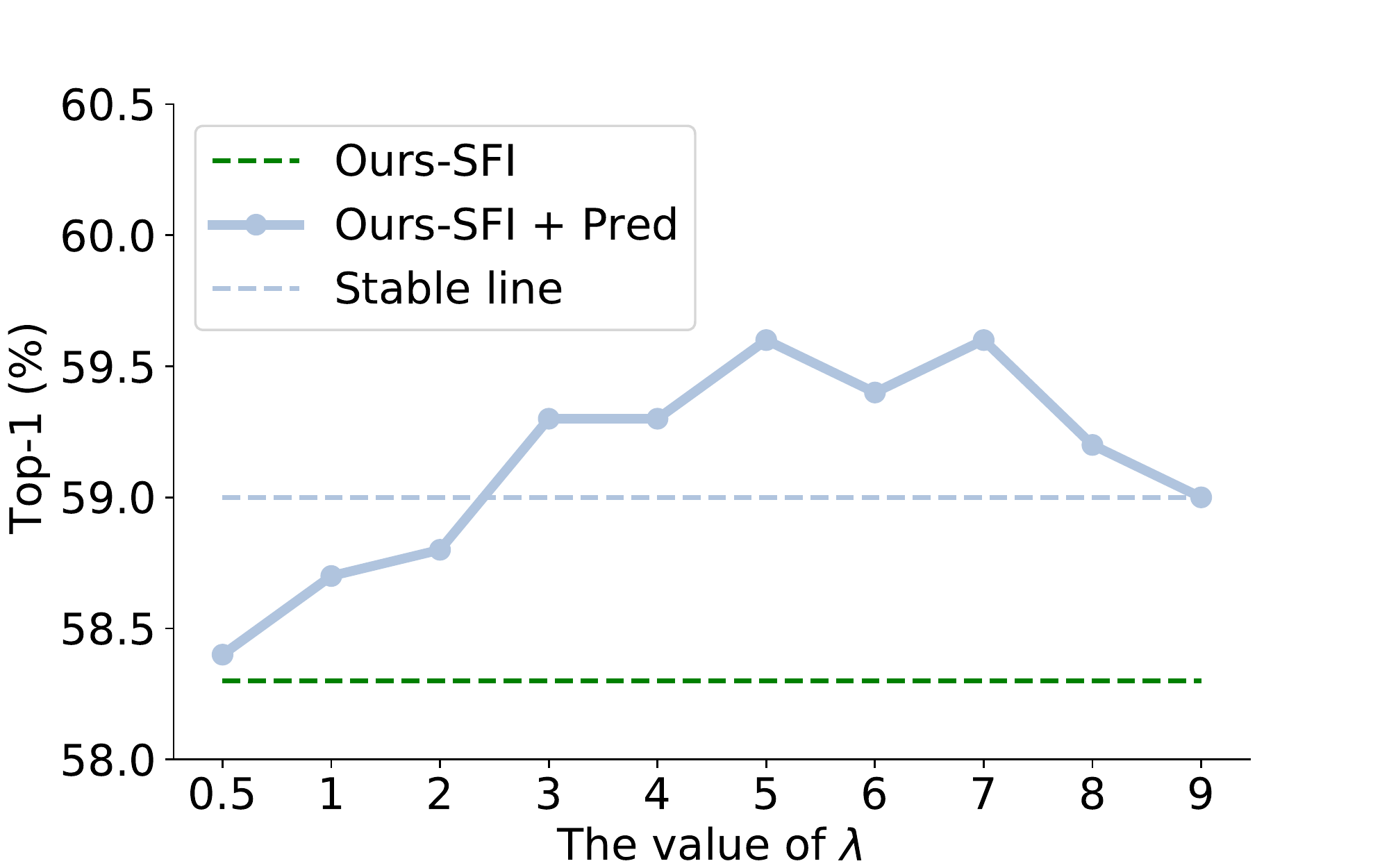}
	\end{center}
	\caption{The effect of $\lambda$ on the Something-Else dataset.}
	\label{fig:lambda}
\end{figure}

\noindent{\bf Effect of $\lambda$ in objective function.} The hyper-parameter $\lambda$ is designed to balance the contribution from the auxiliary prediction task and the main recognition task. {As shown in Fig~\ref{fig:lambda}, Ours-SFI + Pred can achieve stable benefits~(above the dashed blue line) from the auxiliary prediction task when $3 \leq \lambda \leq 9$, compared with Ours-SFI. Thus, we simply set $\lambda$ to $5$ for the proposed approach in experiments.}

\begin{figure*}
	\definecolor{green_}{rgb}{0.61, 0.73, 0.38}
	\definecolor{red_}{rgb}{0.75, 0.31, 0.27}
	\definecolor{cyan_}{rgb}{0, 1, 1}
	\definecolor{blue_}{rgb}{0.47, 0.678, 0.82}
	\definecolor{orange_}{rgb}{1, 0.698, 0.43}
	\begin{center}
		\subfigure[]{
			\includegraphics[width=0.9\linewidth]{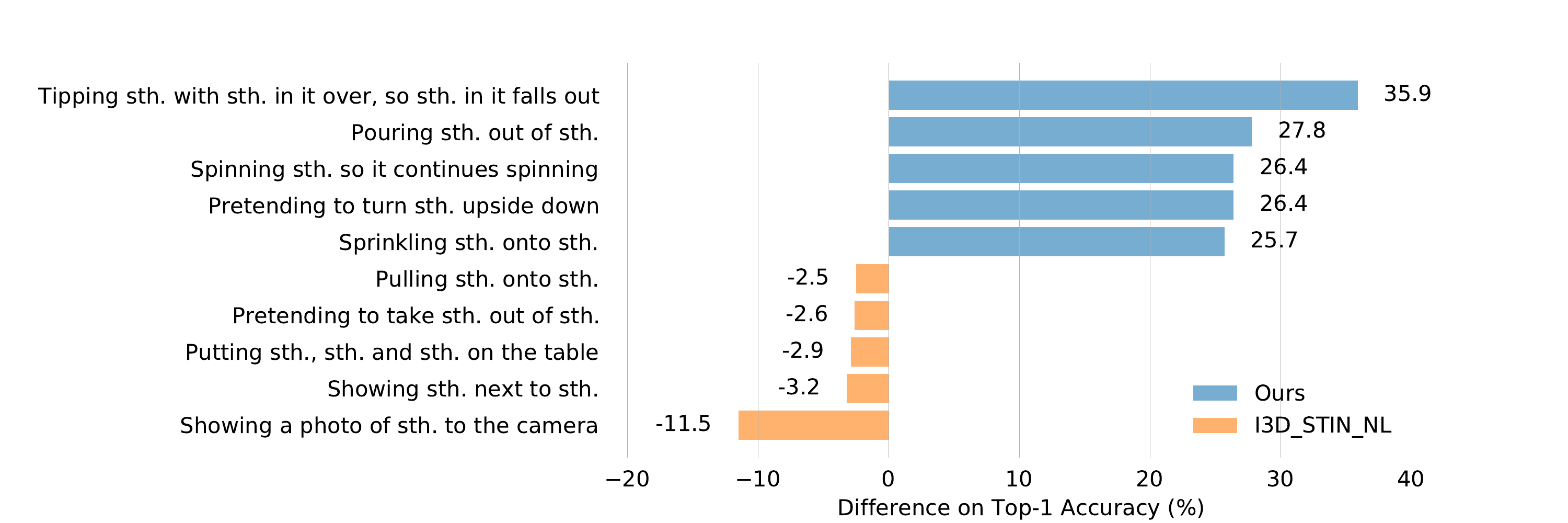}
		}\label{fig:vis_category}
		\subfigure[]{
			\includegraphics[width=0.45\linewidth]{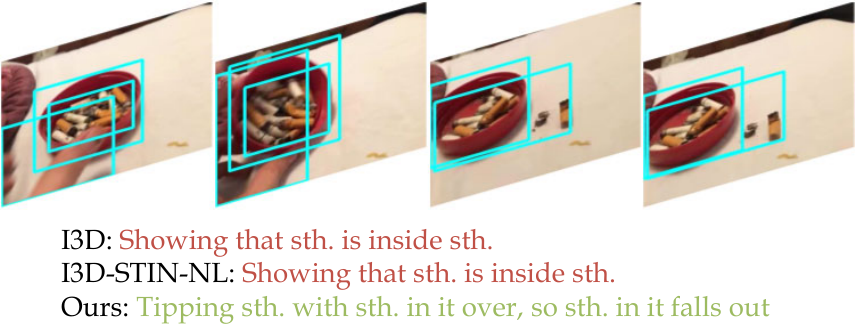}
		}
		\subfigure[]{
			\includegraphics[width=0.45\linewidth]{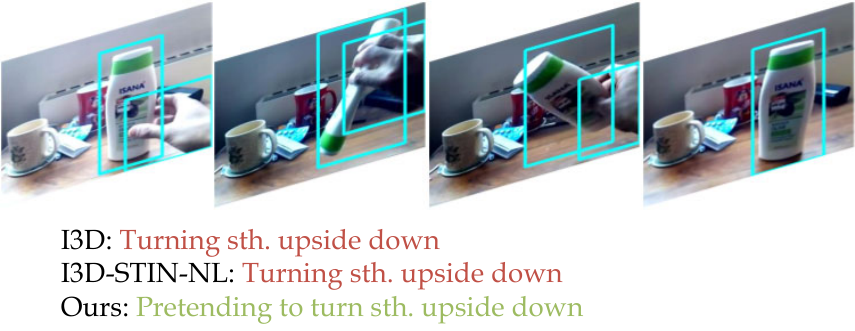}
		}
		\subfigure[]{
			\includegraphics[width=0.45\linewidth]{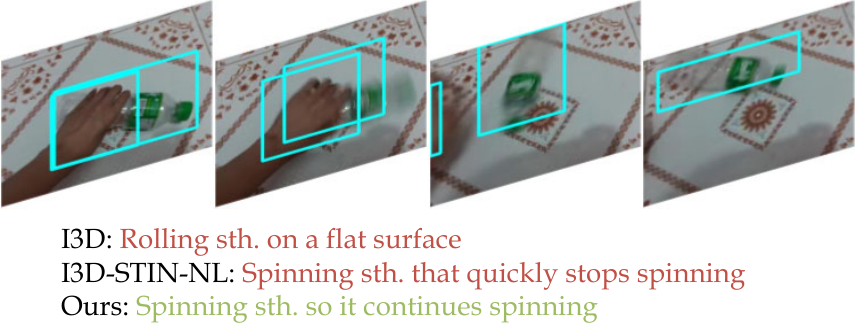}
		}\label{fig:vis_c}
		\subfigure[]{
			\includegraphics[width=0.45\linewidth]{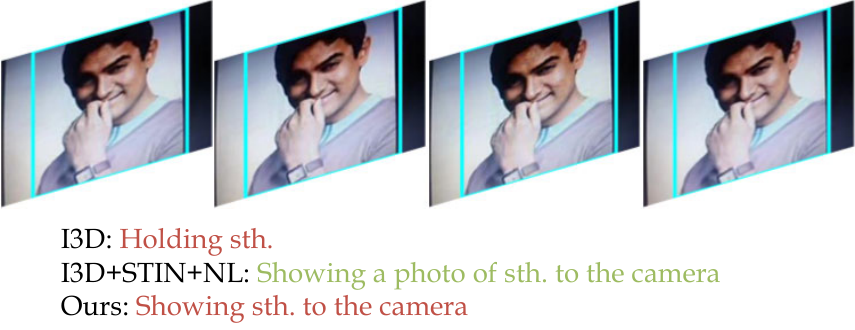}
		}\label{fig:vis_d}
	\end{center}
	\caption{(a): Top-5 categories on which our approach exceeds~({\color{blue_} blue} bar) or lags behind~({\color{orange_}orange} bar) I3D + STIN + NL, respectively. The numbers represent the difference between two models in terms of top-1 accuracy. (b-e): Predictions of I3D~\cite{carreira2017quo, wang2018videos}, I3D + STIN + NL~\cite{materzynska2020something}, and Ours~(\textit{i.e.}, Ours-SFI + Pred) on some examples from the Something-Else dataset. All instances are annotated by {\color{cyan_}cyan} boxes, and the correct and incorrect predictions are highlighted in {\color{green_} green} and {\color{red_}red}, respectively.  Best viewed in color.}
	\label{fig:visualization}
\end{figure*}

\noindent{\bf Category analysis.} Overall, our approach surpasses I3D + STIN + NL~\cite{materzynska2020something} on $\bm {84.5}\%$ categories and the top categories are shown in Figure~\ref{fig:visualization}~(a). In particular, our approach is far beyond~\cite{materzynska2020something} on some categories, such as ``pouring sth. out of sth.", ``spinning sth. so it continues spinning", etc., which involve not only relative motions but also object properties. It is understandable that our method is worse than I3D + STIN + NL on ``showing a photo of sth. to the camera"~(in Figure~\ref{fig:visualization}~(e)) in which only a static photo in the video.

\noindent{\bf Visualization.}~{We also visualize some predictions on the Something-Else dataset in Figure~\ref{fig:visualization}. We can see that I3D + STIN + NL is easy to learn the bad appearance or positional bias due to the rough late fusion of different features. For instance, in Figure~\ref{fig:visualization}~(b), I3D + STIN + NL prefers to capturing the concept of ``sth. is inside sth." when sees a container containing certain objects similar to the I3D model, but ignores the negligible yet very important positional self-change of objects. Besides, I3D + STIN + NL ignores the concept of ``pretend" in Figure~\ref{fig:visualization}~(c) due to the worse bias from positional features. However, our approach generates predictions by fusing different levels of features in an interactive way, rather than prefer to one of them.
}

\section{Conclusion}
{In this work, we first point out that main challenge of compositional action recognition is to integrate features that often differ significantly in modality and dimensionality. To alleviate this problem, we propose a novel \textbf{interactive fusion} framework to project these features across different spaces and promote the fusion process by an auxiliary prediction task. Our framework is composed of three steps, namely, positional-to-appearance feature extraction, semantic feature interaction and semantic-to-positional prediction. Experimental results on two action recognition datasets show the robust generalization ability of our interactive fusion framework on compositional action recognition. 
	This work delivers that the understanding of human action is actually a process of gradually fusing features from multiple sources with diverse strategies. We hope that this work can inspire future research on compositional activity understanding.
}

{\small
\bibliographystyle{ieee_fullname}
\bibliography{egbib}
}

\end{document}